\documentclass{article} 

\usepackage[utf8]{inputenc} %
\usepackage[T1]{fontenc}    %
\usepackage{hyperref}       %
\usepackage{url}            %
\usepackage{booktabs}       %
\usepackage{amsfonts}       %
\usepackage{nicefrac}       %
\usepackage{microtype}      %
\usepackage{fullpage}
\usepackage{lineno}
\usepackage{natbib}
\usepackage{pifont}
\usepackage{amsmath,amssymb}
\usepackage{graphicx}
\usepackage{subcaption}
\usepackage{multicol,graphicx,xcolor,multirow, array}
\usepackage[normalem]{ulem}
\usepackage{mathtools}
\usepackage{lineno}
\usepackage{accents} %
\usepackage[export]{adjustbox} %
\usepackage{diagbox}
\usepackage{gensymb,stmaryrd,mathtools,textcomp,xspace,grffile}
\usepackage{iclr2020_conference,times}
\usepackage{makecell}
\usepackage{adjustbox}

\def \mytitle{Deeply Uncertain: Comparing Methods of Uncertainty Quantification in Deep Learning Algorithms}

\usepackage{amsmath,amsfonts,bm}

\def\eqref#1{eq.~(\ref{#1})}

\def\1{\bm{1}}

\DeclareMathAlphabet{\mathsfit}{\encodingdefault}{\sfdefault}{m}{sl}
\SetMathAlphabet{\mathsfit}{bold}{\encodingdefault}{\sfdefault}{bx}{n}

\def \figcalibrationchangeell
{
\begin{figure}
   \centering %
   \includegraphics[width=\textwidth,keepaspectratio]{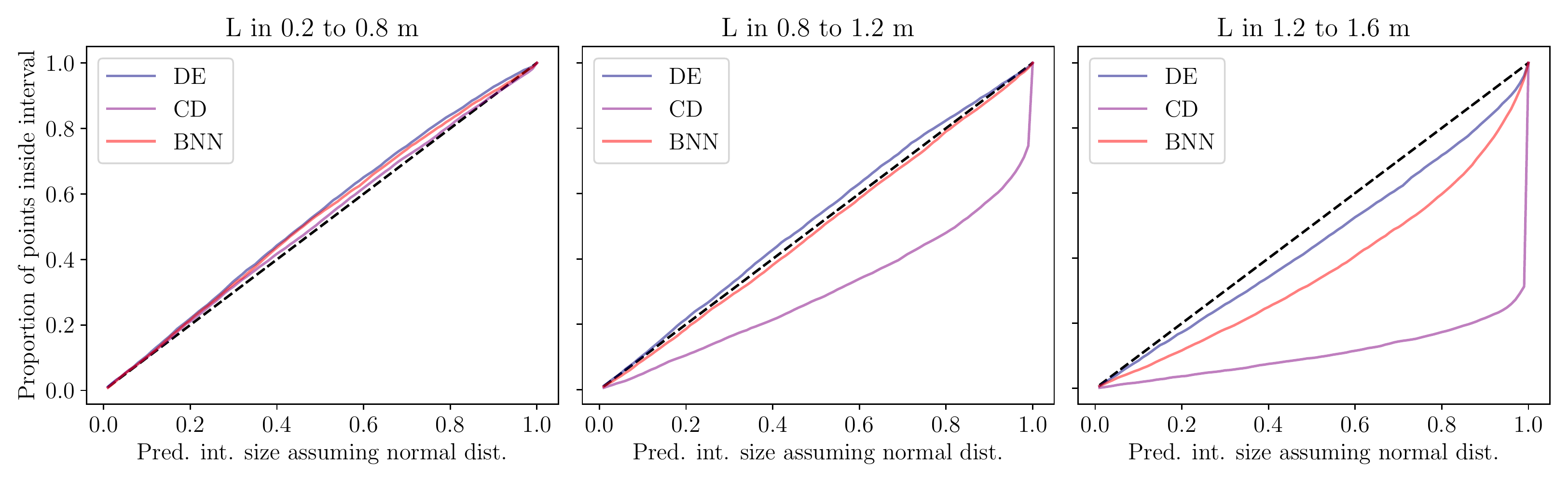}
      \caption{
      Calibration curves for all UQ methods.
      The horizontal axis denotes the fraction of the prediction interval derived assuming a normal distribution of residuals with standard deviation equal to the predictive uncertainty.
      The vertical axis denotes the derived fraction of points that fall within that prediction interval.
      The dashed line (1-to-1) denotes perfect coverage.
      Left: performance of methods in the range where they were trained, $L \in (0.2, 0.8)$ m.
      Middle: performance of methods slightly out of training range, $L \in (0.8, 1.2)$ m.
      Right: performance of methods farther out of training range, $L \in (1.2, 1.6)$ m.
      Within the training distribution, the calibration is not far from the 1-1 line for all methods in the training distribution. 
      Moving rightward, all methods increasingly underestimate the error.
      }
         \label{fig:calibration}
   \end{figure}
}

\def \figood
{
\begin{figure}
\begin{subfigure}{0.48\textwidth}
   \centering %
   \includegraphics[width=\textwidth,keepaspectratio]{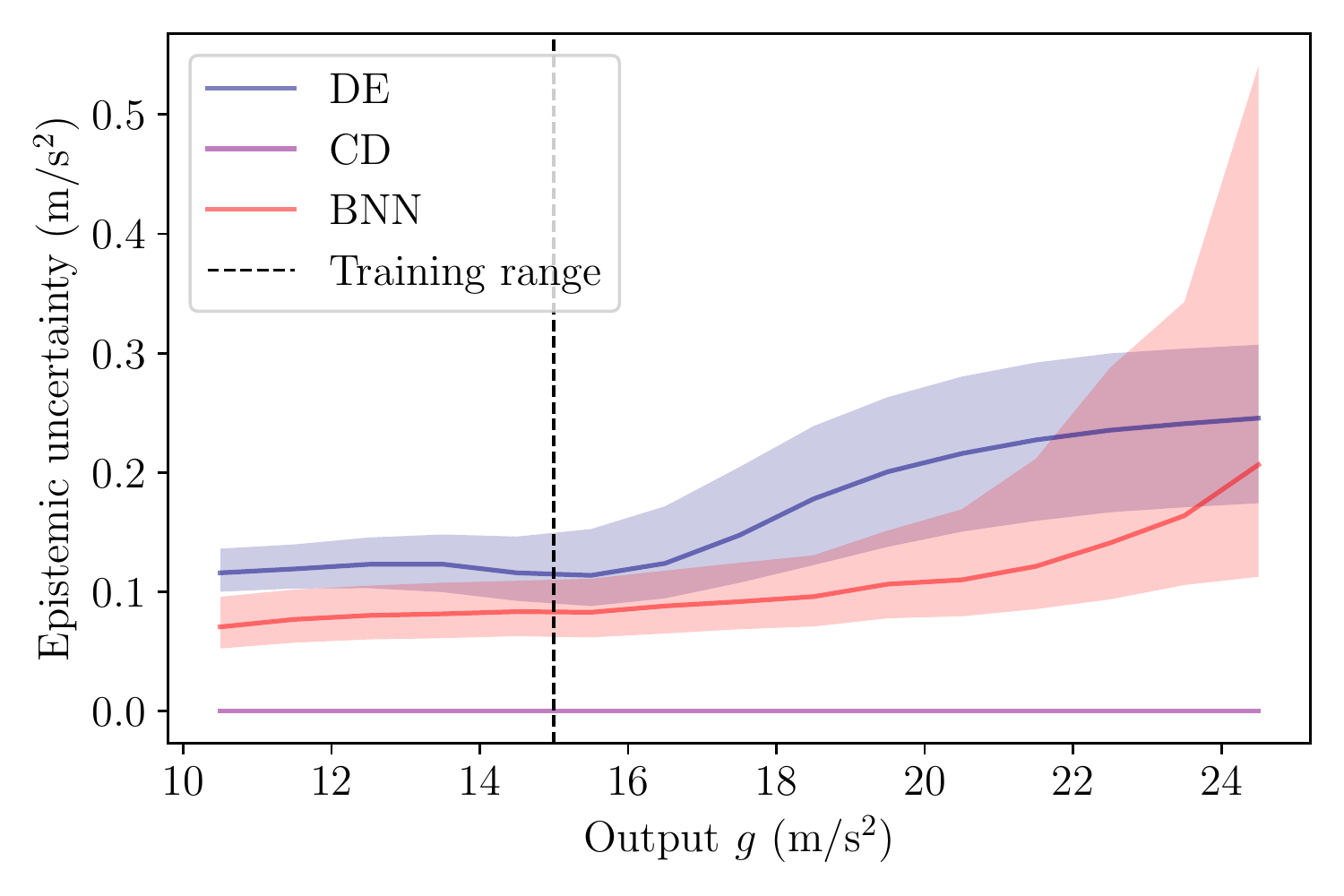}
      \caption{
      Test values of $g$ move away from the network training range, and $L$ remains in range.
      For the training set, $g \in (5,15)$ m/s$^2$.
      }
         \label{fig:oodg_epistemic_unc}
\end{subfigure} \hfill
\begin{subfigure}{0.48\textwidth}
   \centering %
   \includegraphics[width=\textwidth,keepaspectratio]{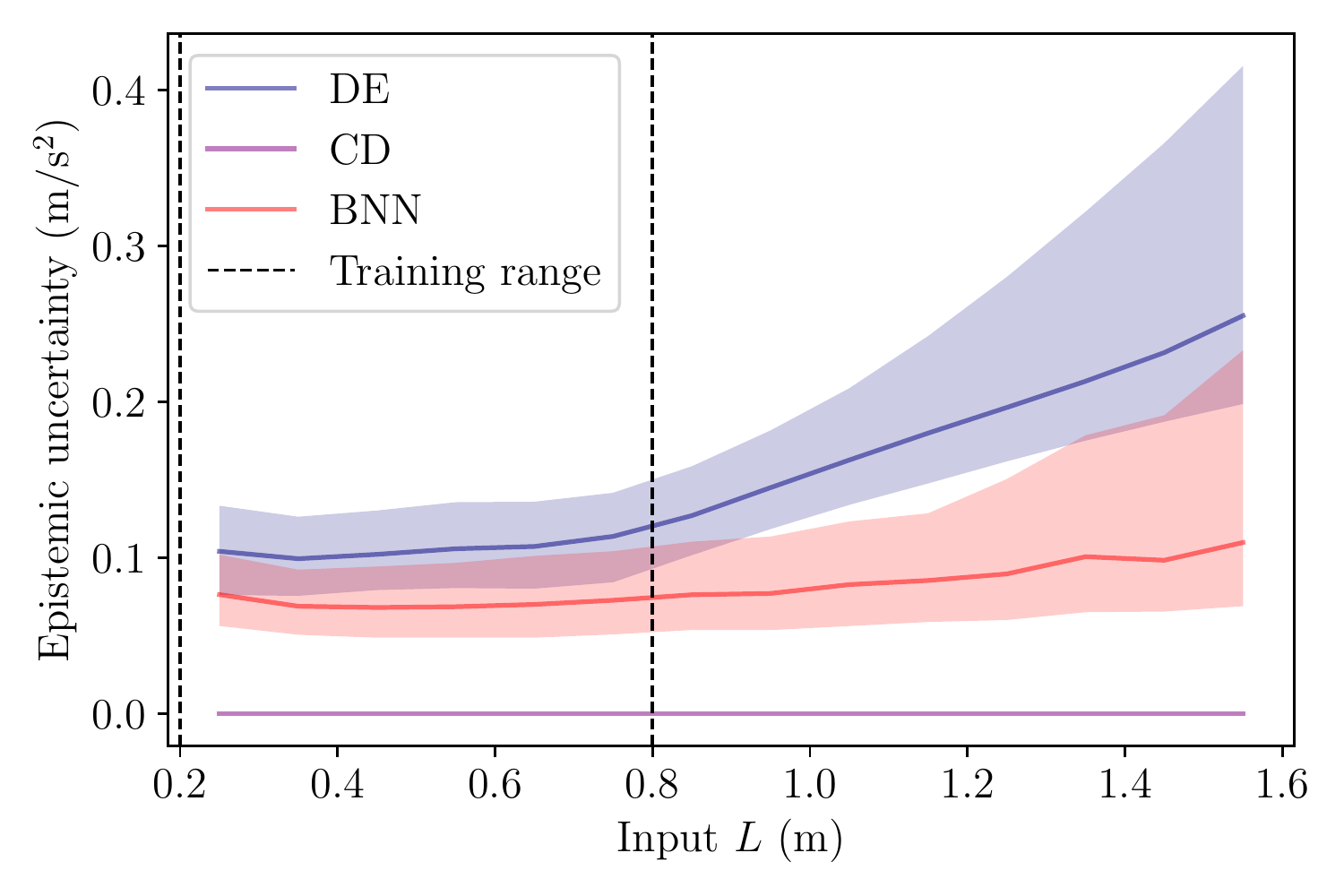}
      \caption{
      Test values of $L$ move away from the network training range, and $g$ remains in range.
      For this training set, $L \in (0.2, 0.8)$ m.
      }
         \label{fig:oodl_epistemic_unc}
\end{subfigure} 
\caption{
      Epistemic uncertainty for various values of the output prediction $g$ and the input value $L$. 
      The dashed line shows the training range for the given parameter.
      As the test data moves away from the training range, epistemic uncertainty estimates rise slowly both for DE (blue) and BNN (red), potentially enabling a classifier to flag out-of-distribution inputs.
      In this experiment, CD (purple) always gave zero epistemic uncertainty for a wide majority of inputs.
      To make this plot, the $x$-axis was divided into equal-sized bins.
      The lines correspond to the median prediction in each bin, while the shaded area denotes the interval between the 16th and 84th percentiles.
      }
         \label{fig:ood}
\end{figure}
}

\def \figcomparisonsnoise
{
\begin{figure}
   \centering %
   \includegraphics[width=\textwidth,keepaspectratio]{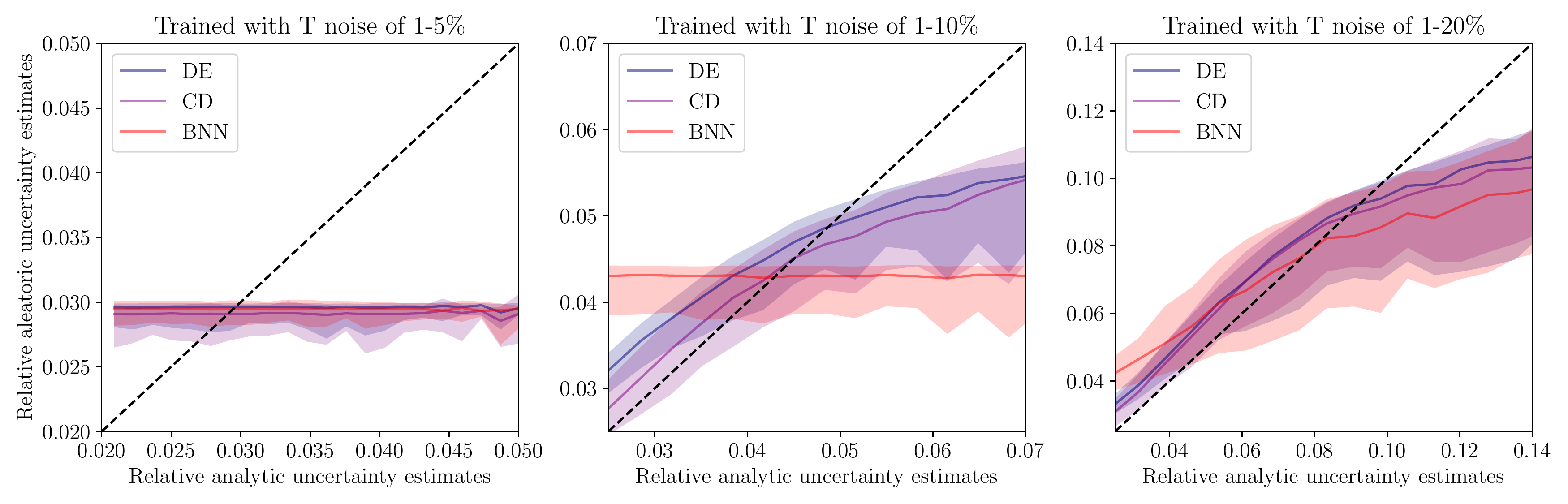}
      \caption{
      Comparison of the relative aleatoric statistical uncertainty in $g$ to the relative analytic uncertainty estimate of $g$ for each method, with increasing ranges of noise in $T$.
      The relative uncertainty is the ratio of the uncertainty and the value of $g$ predicted.
      To make this plot, the $x$-axis was divided into equal-sized bins.
      The lines correspond to the median prediction in each bin, while the shaded area denotes the interval between the 16th and 84th percentiles.
      For the narrowest range of noise in training (left), the deep learning-based UQ methods fail to correlate with the analytic method: they reproduce average noise in the training set, but not the variation. 
      For larger values of noise in $T$ (middle), DE and CD correlate better with the analytic estimate, but they overestimate low analytic uncertainties, and underestimate high analytic uncertainties. 
      BNNs still fail to correlate with the analytic estimate.
      For much larger values of noise in $T$ (right): all methods follow a similar trend to the analytic estimate.
      This flags an issue that one should be aware of when training deep learning methods with a negative log-likelihood objective.
      }
         \label{fig:comparisons_noise}
   \end{figure}
}

\def \tbluncertaintycalibration
{
\begin{table}
        \begin{adjustbox}{center}
        \begin{tabular}{ l c c c c c c c c}
        \hline
        & \multicolumn{3}{c}{Correlation to analytic, $T$ noise:} & &  \multicolumn{3}{c}{Miscalibration, $L$ range:} & MSE \\
        \cline{2-4} \cline{6-8} 
         & 1-5\% & 1-10\% & 1-20\% &  & (0.2, 0.8) m & (0.8, 1.2) m & (1.2, 1.6) m & \\
        DE & 0.0015(3) & \textbf{0.810(6)} & \textbf{0.838(3)} &  & \textbf{0.026(3)} & \textbf{0.0170(18)} & \textbf{0.035(8)} & \textbf{0.4907(2)} \\
        CD & \textbf{0.08(4)} & 0.74(5) & 0.822(11) &  & \textbf{0.019(4)} & 0.07(3) & 0.21(6) & 0.5014(19) \\
        BNN & 0.003(5) & $-$0.0143(19) & 0.72(3) &  & 0.035(3) & 0.026(6) & 0.067(19) & 0.5018(15) \\
        \hline
        \end{tabular} 
        \end{adjustbox}
        \caption{In this table we report the mean results of 6 independent runs along with the uncertainties on the last significant digits of those means.
        On the first three columns, we evaluate how correlated the aleatoric uncertainty estimates are with the analytic uncertainty estimate, in all situations studied in Fig.~\ref{fig:comparisons_noise}.
        When the aleatoric uncertainty estimate is constant, we obtain an approximately null correlation coefficient.
        In all other situations, we find roughly comparable results, with DE edging the other methods.
        Then we evaluate the calibration of each method in the three situations described in the text.
        Miscalibration is defined as the area between the 1-1 line and the calibration lines in Fig.~\ref{fig:calibration}.
        Perfect calibration corresponds then to 0, while 0.5 is the worst possible value.
        As we saw in the graphs, CD is slightly better calibrated in training distribution, but its performance quickly degrades as we move from it.
        Farther from the training distribution, DE clearly beats the other methods.
        Though we did not focus on prediction quality in this manuscript, we present also the mean squared error in the in-training-distribution test set.
        The performance of all methods is close, but DE once again is the winner.
        We highlight the best performance in each column.
        }
        \label{tbl:uncertaintycalibration}
\end{table}
}

\def \tbluncertaintytypesnew{
\begin{table*}[t]
   \centering
   \noindent\begin{minipage}[b]{0.99\textwidth}
   \centering
  \centering
  \begin{tabular}{| l | c c | }
\hline
\diagbox[width=65pt]{Physics}{ML} & Aleatoric      & Epistemic   \\
\hline
Statistical      &  
\makecell{\\ noise in data, \\ stdev of measurements \\ (noise in period $T$)}   & ---      \\
Systematic     &  \makecell{ \\ noise in data, \\ not stdev of measurements \\ (noise in length $L$) }  & \makecell{ \\ model fidelity, \\ not stdev of measurements \\ (far from training set) } \\
\hline
\end{tabular}
\caption{Types of uncertainties in the cross-over between machine learning (ML in the columns) and physics (rows).
    Examples specific to the experiment done here are shown in parentheses.
    \label{table:uncertaintytypes}
    }
\end{minipage}
\end{table*} 
}

\title{\mytitle}

\author{Jo\~ao Caldeira\\
Fermi National Accelerator Laboratory, P.O. Box 500, Batavia, IL 60510, USA \\
\texttt{caldeira@fnal.gov}
\AND
Brian Nord \\
Fermi National Accelerator Laboratory, P.O. Box 500, Batavia, IL 60510, USA \\
Kavli Institute for Cosmological Physics, University of Chicago, Chicago, IL 60637, USA \\
Department of Astronomy and Astrophysics, University of Chicago, Chicago, IL 60637, USA \\
\texttt{nord@fnal.gov}
}

\iclrfinalcopy

\begin{document}

\maketitle

\begin{abstract}
We present a comparison of methods for uncertainty quantification (UQ) in deep learning algorithms in the context of a simple physical system.
Three of the most common uncertainty quantification methods --- Bayesian Neural Networks (BNN), Concrete Dropout (CD), and Deep Ensembles (DE) --- are compared to the standard analytic error propagation.
We discuss this comparison in terms endemic to both machine learning (``epistemic'' and ``aleatoric'') and the physical sciences (``statistical'' and ``systematic'').
The comparisons are presented in terms of simulated experimental measurements of a single pendulum --- a prototypical physical system for studying measurement and analysis techniques.
Our results highlight some pitfalls that may occur when using these UQ methods.
For example, when the variation of noise in the training set is small, all methods predicted the same relative uncertainty independently of the inputs.
This issue is particularly hard to avoid in BNN.
On the other hand, when the test set contains samples far from the training distribution, we found that no methods sufficiently increased the uncertainties associated to their predictions.
This problem was particularly clear for CD.
In light of these results, we make some recommendations for usage and interpretation of UQ methods.

\end{abstract}

\section{Introduction}
\label{sec:introduction}

Methods in machine learning -- and more specifically deep learning -- are increasingly utilized in modeling, analysis, and control procedures in the physical sciences \citep[for a review, see e.g.,][]{Mehta19}.
In those fields, it is critical for every result to be accompanied by a quantification of the uncertainty.
Therefore, confidence in uncertainties reported for machine learning model predictions is a necessity before a widespread adoption of these novel methods by the community.
Even though many uncertainty quantification (UQ) methods for deep learning have been put forth in recent years, uncertainties are often presented in terms of concepts unfamiliar to those working in the physical sciences, such as ``epistemic'' and ``aleatoric'' uncertainties.
In addition, while there have been a number of comparisons between the results of different methods, they are often presented in the context of a very technical task or one that has complex data structures, making generalizations to other tasks less accessible \citep[e.g.][]{Hortua19,Snoek19,Scalia19,Tran19}.

We aim this work to serve as part of a bridge between the scientific communities which use these algorithms as tools and the statistics and computer science communities developing the algorithms.
We present a comparison of multiple UQ methods in a relatively simple physics setup --- a single pendulum experiment.
The regression task undertaken is a calculation that is typically asked of a student in the setting of an undergraduate physics laboratory classroom --- given enough measurements to characterize the motion of a pendulum, calculate the gravitational acceleration $g$.

A crucial element in our setup is the ability to introduce different types of noise in the input data and to propagate that noise into an uncertainty in the resulting prediction that is independent of the method used.
We then observe the results from UQ methods and assess how they reflect the different sources of uncertainty that are injected into the generated data.
We also aim to present the methods in a framing that is readily accessible for practitioners in the physical sciences.
While we aim to set the stage for more widespread usage in science of the UQ methods summarized here, we do not claim to present a definitive comparison of those methods directly applicable to all situations.

\section{Methods: Experimental Setup and Uncertainty Analysis}
\label{sec:methods}

Below, we describe the physical system for the computational experiment, metrics for UQ in both the machine learning and physical sciences domains, and the methods of UQ in deep learning that are analyzed in this work.

\subsection{Uncertainty Quantification Metrics: Aleatoric and Epistemic; Statistical and Systematic}
\label{sec:uncertainty_definitions}

First, we need to create a mapping between different conceptualizations or descriptions of uncertainty.
In machine learning and statistical contexts, \emph{aleatoric} uncertainty originates in corruptions of the input data, such as detector noise or the point spread function in an astrophysical context.
Regardless of the quality of a model or the quantity of training data, this uncertainty can never be evaded or reduced.
\emph{Epistemic} uncertainties describe the fidelity of the model in its representation of the data --- barring aleatoric uncertainties.
Epistemic uncertainties decrease as the training data size increases.
In fields like decision analysis, these two types of uncertainty are called ``risk'' and ``uncertainty,'' respectively.
Longer reviews of these concepts can be found in \cite{Gal16}, section 1.2; \cite{Hullermeier2019}.

Precise definitions of what physicists refer to as ``statistical'' and ``systematic'' uncertainty are somewhat elusive, and there appears to be no definitive community-wide standard in the literature.
We follow conventions from \cite{metrology-guide}, which advocates for the terminology of ``Type A'' (also known as ``statistical'') and ``Type B'' (also known as ``systematic'') uncertainties.
In this work we will use the terms statistical and systematic, as they are more common.
\emph{Statistical} uncertainty describes errors that can be quantified by statistical analysis of a series of experimental measurements --- i.e., a standard deviation of a random variable sampled by repeated measurements under the same conditions.
\emph{Systematic} uncertainties are quantified by any other means --- e.g., originating in a model, drawn from statistical uncertainties in measurements performed by a different experiment, and theoretical uncertainties.
Systematic uncertainties include those which cannot be reduced by an increase in data from the same experiment.
It should be noted that a systematic uncertainty here still corresponds to a standard deviation on the distribution of a measurement value.
This is distinct --- and does not include --- a systematic bias making the measurement consistently too high or too low which can be corrected before quoting a final measurement value.

From the definitions above, we summarize that a) epistemic errors are always systematic, and statistical errors are always aleatoric and never epistemic; and b) aleatoric uncertainties can be systematic.
There are then three types of relevant uncertainties: \emph{aleatoric systematic}, \emph{aleatoric statistical}, and \emph{epistemic systematic}.
We will provide explicit examples of these uncertainties in the context of the pendulum in section~\ref{sec:metrics}.

\subsection{Physical System: Single Pendulum}
\label{sec:pendulum}

The physical system for this experiment was chosen to meet the following qualifications:
\begin{enumerate}
    \item The physical processes, calculations, and sources of uncertainty  should be readily understandable by any physicist or machine learning scientist.
    \item There are statistical and systematic sources of uncertainty, and we can quantify them for each example to compare the results to those of the machine learning algorithms.
\end{enumerate}
An appealing scenario that meets these requirements is the single-pendulum experiment, which is common in an undergraduate physics laboratory setting and is analytically well modeled by 
\begin{equation}
    g = 4 \pi^2 \frac{L}{T^2}, \label{eqn:pendulum_eqn}
\end{equation}
where $g$ is the gravitational acceleration, $L$ is the length of the pendulum arm, and $T$ is the period of oscillation.
For each pendulum instance, we simulate mock measurements like those one might take in a pendulum experiment, including a mass $m$ for the pendulum, the maximum oscillation angle $\theta$, length of the pendulum arm $L$, and ten independent measurements of the period $T$.
In summary, we will train machine learning models to model the acceleration $g$ from 13 inputs.

Note that $g$ is independent of $m$ and $\theta$, but we include them in the inputs in order to simulate the full set of measurements one might make in a laboratory setting.
Though we will not do so here, these could be used to introduce additional systematic uncertainties.
For instance, we train the model using \eqref{eqn:pendulum_eqn}, which is valid only for small $\theta$.
Moving away from the range of validity of \eqref{eqn:pendulum_eqn} should come with an increased uncertainty in the result.

\subsection{Sources of Uncertainty and Metrics for Comparison}
\label{sec:metrics}
In our computational experiments, we seek to distinguish how types of uncertainty quantification are reflected in the outputs of each model.
In section~\ref{sec:uncertainty_definitions} we outlined three different types of uncertainty.
We can inject each of them into the data-generation process outlined in section~\ref{sec:pendulum} as follows.
\begin{itemize}
    \item \emph{Aleatoric statistical} uncertainty can be included by adding noise in the 10 measurements of the period, $T$.
    For each data point in the training set, we draw the amount of measurement noise $\nu$ uniformly in some range, and then draw each measurement of the period from a normal distribution with standard deviation $\nu T$.
    The choice of the range for $\nu$ in the training set merits a longer discussion in section \ref{sec:results}.
    \item \emph{Aleatoric systematic} uncertainty exists if the single measurement of $L$ also contains noise, as this is a source of uncertainty that cannot be statistically determined from the single measurement of $L$.
    Note that since there is no statistical way to determine this noise from the input data alone, the uncertainty must be determined from the typical noise seen in training.
    In our training and test sets, all measurements of $L$ are drawn from a normal distribution with standard deviation $0.02 L$.
    \item \emph{Epistemic systematic} uncertainty reflects how uncertain the model is of its predictions.
    One way to test this is by looking at predictions far from the training set manifold.
    In this experiment, we train networks with $g \in (5, 15)$ m/s$^2$, and $L \in (0.2, 0.8)$ m.
    Either of these can be moved outside that range, and we will consider both cases below.
\end{itemize}

We investigate several ways of comparing different UQ methods.
For the two aleatoric uncertainty sources, we calculate the uncertainty in the output analytically using standard uncertainty propagation.
This means that we use \eqref{eqn:pendulum_eqn} to propagate the effect of noise in each input variable to the acceleration $g$.
We then compare the aleatoric uncertainties found by the machine learning algorithms to the analytic estimate.
This permits a direct comparison with which to examine deep learning methods: standard uncertainty propagation produces an uncertainty that is considered ``truth.''\footnote{Strictly, the uncertainty estimate used here is a linear approximation. The error in that approximation is small in the ranges considered in this work.}
For the epistemic uncertainty, we do not perform an analytic estimate, but investigate how this deep learning uncertainty changes as the test data move away from the training distribution.
The expectation is that it should increase with distance of the test data from the training data.
Finally, we analyze the prediction intervals from each deep learning-based method: the reliability diagram (also called ``calibration curve'') compares the proportion of samples that fall within a prediction interval to the expectation for that proportion if the uncertainty represents a standard deviation of a normal distribution.

Please see Table~\ref{table:uncertaintytypes} for a concise summary of the correspondences in these uncertainty conceptualizations, as well as examples for this particular physics model.

\tbluncertaintytypesnew

\subsection{Uncertainty Quantification Methods for Deep Learning Models}
\label{sec:uqmethods}

We will include three uncertainty quantification methods: Deep Ensembles (DE), Bayesian Neural Networks (BNN), and Concrete Dropout (CD).
For all these methods, epistemic uncertainty is estimated by looking at an ensemble of trained models, though the method of sampling from the set of possible models varies from method to method.
The spread of predictions between different models is used as an estimate of model-related (i.e., epistemic) uncertainty.
Aleatoric uncertainty, on the other hand, can be predicted by a single model as it is related to the amount of observation noise in a given region of the input space.
The effect of that noise on the result is estimated by fitting both the mean and standard deviation of a normal distribution to maximize the log likelihood of the data.
The standard deviation obtained is an estimate of aleatoric uncertainty.
A review of this distinction can be found in \cite{Kendall17}.

For each model and experiment, then, we will obtain $N=10$ estimates of the prediction mean and aleatoric uncertainty, $(\mu_i, \sigma_i)$.
We combine the $N$ estimates as a mixture of Gaussians, and obtain the following predictions:
\begin{align}
\hat{g} &= \frac{1}{N} \sum_{i=1}^N \mu_i  = \text{mean}(\mu_i) && \text{(gravitational constant mean)} \\
\sigma_{al} &= \sqrt{\frac{1}{N}\sum_{i=1}^N \sigma_i^2} = \sqrt{\text{mean}(\sigma_i^2)} && \text{(aleatoric uncertainty)} \\
\sigma_{ep} &= \sqrt{\frac{1}{N}\sum_{i=1}^N \mu_i^2 - \hat{g}^2}  = \text{stdev}(\mu_i) && \text{(epistemic uncertainty)}\\
\sigma_{pr} &= \sqrt{\sigma_{al}^2+\sigma_{ep}^2} && \text{(total predictive uncertainty)} 
\end{align}

\subsubsection{Bayesian Neural Networks}
Bayesian Neural Networks (BNN) are a class of neural networks in which the weights of each layer form a valid probability distribution \citep{Graves11}.
The training process then consists of approximate Bayesian inference on these probability distributions, given the data.
Exact inference on the network parameters is an intractable problem, we approximate it by the evidence lower bound (ELBO).
In practice, this consists of a sum of the negative log likelihood of the data with the Kullback-Leibler (KL) divergence between the weight distribution and the weight prior.

There have been several methods proposed to efficiently sample from the weight distributions in the training process while keeping the parallelization techniques available to usual neural networks.
In this work, we make use of flipout \citep{Wen18}, as implemented in the TensorFlow Probability library \citep{Dillon17}.
Following \cite{Snoek19}, we found a large improvement in predictions after scaling the KL divergence term in the loss by $1/n$, where $n$ is the number of training examples.
We also attempted replacing KL divergence with Maximum Mean Discrepancy, as advocated by \cite{Pomponi20}.
The results did not change appreciably, so we use KL in the experiments shown.
With BNN, we can evaluate epistemic uncertainties by looking at the different outputs produced when we sample multiple times from the posterior weight distributions.

\subsubsection{Deep Ensembles}
Deep ensembles (DE), introduced as a simpler alternative to Bayesian methods in \cite{Lakshminarayanan16}, are attractive because of their conceptual simplicity: we simply need to retrain the same network many times with different initializations.
The randomness inherent in the initializations and in the training process then provides different samples of trained network parameters.
If we optimize the networks to minimize the mean squared error loss, this provides only a measure of epistemic uncertainty.
On the other hand, as outlined above, we can optimize the data log likelihood and then we estimate both aleatoric and epistemic uncertainties.
A related technique is \emph{bagging}, short for ``bootstrap aggregating.''
This adds another source of randomness by training each network with a different random draw with replacement from the training set.
\cite{Lakshminarayanan16} observes a performance deterioration when using bagging, and we have seen no improvement when bagging is added.

\subsubsection{Concrete Dropout}
Dropout \citep{Hinton12} was first introduced as a form of regularization in neural networks.
The technique consists of omitting a certain percentage of neurons at each layer, with the omitted neurons chosen at random for each pass.
As a regularization technique, dropout is only used during training.
It was later understood \citep{Gal16paper} that keeping dropout during the testing phase could also provide a way to obtain a distribution of possible models.
Methods were then developed to optimize the dropout probability at each layer during training \citep{Gal17}, and this technique is called Concrete Dropout (CD).
To estimate epistemic uncertainties with CD, we simply drop a different set of neurons on each pass and look at the distribution of results obtained.

\subsection{Network Architecture and Training}
For all methods, we train fully-connected networks with three hidden layers and 100 nodes on each hidden layer.
A ReLU activation function is used in the hidden layers, and identity activation function on the final layer.
As the structure of the neural network is not changed, the epistemic uncertainty here does not include how predictions may vary if we change the architecture.
The contribution of such a term should be negligible in this problem, as the models utilized have sufficient capacity to reach an optimal solution.

We use the Adam optimizer \citep{Kingma14} with learning rate $10^{-3}$ for all models except BNN, where a lower learning rate of $10^{-4}$ was necessary for convergence.
Networks are trained for 200 epochs in the case of CD and BNN, and 40 epochs for DE, on 90000 training points.
All networks were implemented on TensorFlow 2 \citep{tensorflow2015-whitepaper} and trained on a desktop with two RTX 2080 Ti GPUs.
Training of each model takes roughly 100 minutes for all UQ methods considered.
The results presented in our plots refer to the results of a single run, but six full runs were made and the results did not change qualitatively.
Table \ref{tbl:uncertaintycalibration} summarizes the results of all six runs.

All code necessary to reproduce the results of this project, and run more experiments using the same data generator, can be found at \url{https://github.com/deepskies/DeeplyUncertain-Public}.

\section{Results}
\label{sec:results}

Recall that we insert Gaussian noise in the measurements of the period $T$, with the relative amount of noise for each point drawn uniformly from a range.
From the spread of $T$ measurements, we calculated an analytic estimate of the uncertainty coming from the noise in the data by standard error propagation methods, and compare that estimate to the aleatoric uncertainty predicted by the deep learning UQ methods discussed in section~\ref{sec:uqmethods}.

We performed this for three ranges of noise $T$ and the results are shown in Fig.~\ref{fig:comparisons_noise}.
In our initial experiments, noise was sampled from the range between 1\% and 5\%.
All three UQ methods predict the same relative uncertainty for all points in the test set, independent of the noise in that particular data point.
We then increased the range of noise present in measurements to be sampled between 1\% and 10\%.
Estimates from CD and DE now follow the trend of analytic estimate well.
However, the BNN continued to predict a constant relative uncertainty for all points.
Finally, when we increase the range of noise in $T$ to be between 1\% and 20\%, all methods now follow the trend of the analytic estimates to some degree, as we can see in the right panel of Fig.~\ref{fig:comparisons_noise}.

\figcomparisonsnoise

The shape of these predictions reflects a tendency that is well-known for predictions of machine learning algorithms \citep[e.g.][appendix]{Ntampaka20}, but we believe has not been documented for uncertainty predictions: after an initial stage of training, models will often predict the mean value of the training set independently of inputs.
The model will only then learn to differentiate different data points, but this may never happen if the training set does not contain enough variation.
This failure mode is less visible for uncertainty predictions, since in a typical problem we do not have an analytic estimate of what the uncertainty should be for each point.
It should be noted that some metrics to evaluate UQ methods, such as reliability diagrams, may give very good results for a model that gives a constant relative uncertainty.
In conclusion, this issue may go undetected if care is not taken to avoid it.

All results shown from this point forward will make use of the networks trained on a noise range between 1\% and 20\%.
We note that aleatoric systematic uncertainty was included in the experiments above and well-modeled by all the methods in all experiments, and therefore we always include 2\% Gaussian noise in the measurements of $L$.
It should also be noted that in Fig.~\ref{fig:comparisons_noise}, uncertainty is overestimated for most samples of the left half of the plots, and underestimated on the right.
The trend is also more closely followed on the lower range of relative uncertainties.
We believe this is caused by an unequal sample of relative uncertainty in the training set: our recipe to build the training set creates a distribution of relative uncertainties which is asymmetric with a longer right tail.
This is a common phenomenon of predictions in machine learning.

We next explore epistemic uncertainties by performing experiments with test sets far from the training distribution: the predictions for this uncertainty should increase the farther the input data is from the training distribution. %
This is a minimal reasonable requirement for any epistemic uncertainty estimate, as the model should be more uncertain about its prediction in a region where it was not trained.
This could potentially be used to classify certain inputs as out-of-distribution.

\figood

The results of this experiment are shown in Fig.~\ref{fig:ood}.
The expected trend is present for both DE and BNN to varying degrees.
However, CD epistemic uncertainty is very small for a large majority (though not all) of the points in the test sets presented here, even with increasing distance from the training distribution.
This is because the dropout probabilities at each layer are very small -- smaller than $10^{-4}$ for all layers -- and therefore few if any neurons are dropped at each pass.
Since our training set is large and the task at hand is not very complex, it is not unreasonable for epistemic uncertainties to be very low inside the training manifold.
However, one should be aware that when that is the case, CD may be overconfident as we move away from that manifold.

For DE and BNN, we may ask if the rise in epistemic uncertainty is enough to compensate the predictable decrease in accuracy as we move away from the training set.
As $g$ moves away from the training distribution, all models tested here made predictions near the high-end of that training distribution, but never higher, despite the fact that this is not enforced by the network architecture.
The epistemic uncertainty severely underestimates the magnitude of the errors made as the input moves farther above 15 m/s$^2$.

A somewhat simpler ask is to test if the epistemic uncertainties are accurate as the inputs move far from the training manifold, while keeping the outputs inside the training distribution.
We do this by moving $L$ and $T$ together while $g$ remains between 5 and 15 m/s$^2$.
For this experiment, the reliability diagrams for different ranges of $L$ are shown in Fig.~\ref{fig:calibration}.
In the left panel, we see that when $L$ is in the training range, predictive uncertainties are well-calibrated for all methods presented.
As $L$ moves out of distribution (middle panel), CD predictive uncertainties become grossly underestimated, because the epistemic uncertainty predictions do not rise accordingly.
If $L$ is far enough away from the training distribution, as in the right-hand panel, all methods give underestimated uncertainty predictions.

\figcalibrationchangeell

The results of our experiments are summarized in Table~\ref{tbl:uncertaintycalibration}, which contains the average performance over 6 independent runs on the same data.
We note that when the deep learning models come out of the mode where the relative uncertainty predicted is constant, they all achieve roughly comparable correlations with the analytic uncertainty estimates, though the correlation achieved by DE is higher.

\tbluncertaintycalibration

\section{Conclusion and Outlook}
\label{sec:conclusion}

The results outlined above allow us to make some recommendations for usage of these UQ methods.
In order to obtain an accurate estimate of aleatoric statistical uncertainty, care must be taken to have a wide-enough variation in the noise present in the training set, so the model does not become stuck predicting the same relative uncertainty for all points.
This pitfall was particularly hard to come out of in the case of BNN.
In a typical situation (i.e., where no analytic uncertainty estimates are available), one should always check whether the aleatoric uncertainty estimates vary from point to point.
If they do not, the training set should be augmented with examples including a higher amount of noise.
This can be done by collecting additional data with more noise, or by creating a bigger training set including both the original samples and a modification of those samples with added artificial noise.
This data augmentation technique is also used in contexts unrelated to uncertainty quantification.

Aleatoric systematic uncertainties are well-modeled in our experiments, though once again we must make sure the training set is representative.
For systematic uncertainties, it is important to keep in mind that the model can only infer the typical uncertainty in each region of inputs from the training set, as the uncertainty cannot be statistically derived from the inputs.

For epistemic uncertainties, all methods failed to detect how far the inputs had moved from the training distribution.
Even for methods that give a rising uncertainty estimate as the input moves away from the training set, that value very quickly becomes an underestimate when compared to the errors made by the networks.
In particular, CD converged to a very low dropout probability in training, causing it to predict very low epistemic uncertainties in all situations.
While DE and BNN could be used to detect out-of-distribution examples, their quantitative estimates of epistemic uncertainty are not reliable in that situation.

Putting all the results together, we recommend DE.
DE results are the best or comparable to the best in all tests made here, and DE has the advantage of having the smallest conceptual load: one simply needs to train the network several times.
This agrees with the conclusions of earlier comparisons including DE \citep{Snoek19,Scalia19}, although our set up and variable noise in the training set allow us to make novel recommendations.
In the revision stages of this manuscript, we became aware of new methods developed to create ensembles without requiring the training of multiple networks \citep{Madras19, Ashukha20}.
It would be worthwhile to test them against the benchmark presented here.

It would also be valuable to extend the analysis in this manuscript to different neural networks, namely convolutional neural networks.
An equally natural extension would be to move towards more complex data and sources of uncertainty, better reflecting what is needed for specific applications.
With publication we are releasing the code we used to run the experiments described here.
We hope this simple experiment may provide a valuable testbed and benchmark for deep learning UQ methods.

\section*{Acknowledgments}

We thank the University of Chicago Department of Physics Instructional Labs Staff, particularly D.~McCowan, for discussions on what the right experiment would be, the lending of pendulums, and references on uncertainties. 
We thank S.~Basart for input and participation in early stages of this project, A.~Whaley for help in implementing concrete dropout, and T.~Charnock for tensorflow wizardry. 
We also acknowledge valuable comments from H.~Hurd, M.~Hutchinson,  J.~R.~Mart\'inez-Galarza, M.~Ntampaka, C.~Schafer, and H.~Tak.

\subsection*{Author Contributions}
The authors initiated the project concept together.
J.~Caldeira performed all the computations and bulk of the writing.
B.~Nord contributed ideas to solving a few key problems and contributed to the writing.

This work is supported by the Deep Skies Community (\url{deepskieslab.com}), which helped to bring together the authors and reviewers. 
The authors of this paper have committed themselves to performing this work in an equitable, inclusive, and just environment, and we hold ourselves accountable, believing that the best science is contingent on a good research environment.

This manuscript has been authored by Fermi Research Alliance, LLC under Contract No. DE-AC02-07CH11359 with the U.S. Department of Energy, Office of Science, Office of High Energy Physics.

\bibliography{bibliography}
\bibliographystyle{iclr2020_conference}

\end{document}